\pdfoutput=1
\documentclass[11pt]{article}
\usepackage{refcount}

\usepackage[final]{acl}

\usepackage{times}
\usepackage{latexsym}

\usepackage[T1]{fontenc}
\usepackage[utf8]{inputenc}

\usepackage{microtype}

\usepackage{inconsolata}

\usepackage{graphicx}

\title{Memory-based Language Models: An Efficient, Explainable, and Eco-friendly Approach to Large Language Modeling}

\author{
  \textbf{Antal van den Bosch\textsuperscript{1}},
  \textbf{Ainhoa Risco Pat\'{o}n\textsuperscript{1}},
  \textbf{Teun Buijse\textsuperscript{1}}, \\
  \textbf{Peter Berck\textsuperscript{2}},
  \textbf{Maarten van Gompel\textsuperscript{3}}
\\
  \textsuperscript{1}Utrecht University,
  \textsuperscript{2}Lund University,
  \textsuperscript{3}Royal Netherlands Academy of Arts and Sciences
\\
  \small{
    \textbf{Correspondence:} \href{mailto:a.p.j.vandenbosch}{a.p.j.vandenbosch@uu.nl}
  }
}

\begin{document}
\maketitle
\begin{abstract}
We present memory-based language modeling as an efficient, eco-friendly alternative to deep neural network-based language modeling. It offers log-linearly scalable next-token prediction performance and strong memorization capabilities. Implementing fast approximations of $k$-nearest neighbor classification, memory-based language modeling leaves a relatively small ecological footprint both in training and in inference mode, as it relies fully on CPUs and attains low token latencies. Its internal workings are simple and fully transparent. We compare our implementation of memory-based language modeling, {\sc olifant}, with GPT-2 and GPT-Neo on next-token prediction accuracy, estimated emissions and speeds, and offer some deeper analyses of the model.
\end{abstract}

\section{Introduction}

Memory-based language modeling, proposed by \citet{VandenBosch05b}, is a machine learning approach to next-token prediction based on the $k$-nearest neighbor ($k$-NN) classifier \cite{Aha+91,Daelemans+05}. This non-neural machine learning approach relies on storing all training data in memory, and generalizes from this training data when classifying unseen new data using similarity-based inference. Memory-based language modeling is functionally equivalent to decoder Transformers \cite{10.5555/3295222.3295349,radford2019language}, in the sense that both can run in autoregressive text generation mode and predict (a distribution of) next tokens based on a certain amount of prior context.

While training a memory-based language model is generally low-cost, as it involves a one-pass reading of training data and does not involve any convergence-based iterative training, a naive implementation would render memory-based language modeling useless for inference. The upper-bound complexity of $k$-nearest neighbor classification is notoriously unfavorable, i.e. $O(nd)$, where $n$ is the number of examples in memory, and $d$ the number of features or dimensions (e.g. context size). However, improvements and fast approximations are available. \citet{Daelemans+10} offer a range of approximations offering fast classification and data compression using prefix tries. Another notable aspect of memory-based language modeling, as observed earlier by \citet{van-den-bosch-2006-word} for training sets of tens of millions of words, is that its next-word prediction performance appears to increase log-linearly: with every 10-fold increase in the amount of training data, memory-based next-word prediction accuracy appears to increase by a more or less constant amount. This trend may change with unobserved larger amounts of training data; it may also reach a plateau which we currently do not reach because of memory limitations.

The relatively low costs in learning as well as inference make memory-based language modeling a potential eco-friendly alternative to the generally costly training of Transformer-based language models \cite{strubell-etal-2019-energy}. 
This is what this paper aims to explore and document. All experiments with our implementation of memory-based language modeling, a system called {\sc olifant}\footnote{{\em Olifant} is the Dutch word for elephant. Quoted from \url{https://en.wikipedia.org/wiki/Elephant_cognition}, "Most contemporary ethologists view the elephant as one of the world's most intelligent animals. Elephants manifest a wide variety of behaviors, including those associated with grief, learning, mimicry, playing, altruism, tool use, compassion, cooperation, self-awareness, memory, and communication."}, have been carried out with publicly available software, with TiMBL as the basic classification engine.\footnote{\label{timbl}\url{https://github.com/LanguageMachines/timbl}} All required scripts for training and inference with {\sc olifant} are available on GitHub.\footnote{\url{https://github.com/antalvdb/olifant}} Although pre-trained {\sc olifant} models may be run on standard consumer hardware with normal CPUs and reasonably modest RAM requirements, scaled training does eventually require quite large amounts of RAM. The largest pre-trained {\sc olifant} models currently available require in the order of 256 GB of RAM to run.

The paper is structured as follows. Section~\ref{architecture} sketches the architecture of {\sc olifant} and provides an overview of previous and related work. In Section~\ref{learningcurves} we offer learning curve analyses on the next-token prediction task, including carbon emission estimates, in which we compare the different $k$-NN approximations available, and their effect on token generation speed and accuracy. We also compare against differently sized variants of Transformer-based next-word predictors, GPT-2 and GPT-Neo, of which we broadly know how much data they have been trained on. In Section~\ref{memorization} we focus on a specific property of {\sc olifant} that sets it apart from neural LMs, i.e. its ability to memorize and recite training data.

In the remainder of the paper, we offer additional analyses that highlight some of the extensions of the approach. 
%In Section~\ref{embeddings} we explore static embeddings as an alternative way to represent the prompt. 
In Section~\ref{zipf} we analyse the token distributions generated by {\sc olifant} through a frequency lens to test a hypothesis put forward by \citet{KhandelwalLJZL20}, and show by comparison how the sparse distributions of {\sc olifant} variants lead to lower perplexities on validation data when compared to neural models. We offer some conclusions in Section~\ref{conclusions} and critically assess the current limitations of the approach.

\section{{\sc Olifant}: Background and architecture}
\label{architecture}

We begin with a brief overview of previous and related work. We then describe the architecture used in our three alternative implementations of memory-based language modeling available in {\sc olifant}, ranging from lossless $k$-nearest neighbor classification to lossy decision-tree classification.

\subsection{Previous and related work}

Memory-based language modeling is rooted in two classic machine learning paradigms: $k$-nearest neighbor ($k$-NN) classification \cite{fix+51,cover+67,Aha+91} and decision-tree classification \cite{Quinlan86}. Its internal structure is based on an efficient retrieval-oriented tree-shaped index, a prefix trie\footnote{"Trie" as an amalgam of {\em tree} and re{\em trie}val.} \cite{Knuth73}, in which a training set of classification instances can be stored and retrieved efficiently. Using the concept of information gain \cite{Quinlan86} or its robust variant gain ratio \cite{Quinlan93} as the feature ordering heuristic, \citet{Daelemans+93c} introduced IB1-IG as a feature-weighted implementation of $k$-NN, and \citet{Daelemans+97} introduced IGTree as a decision-tree classifier, both making use of on Knuth's prefix tries. Both are implemented in TiMBL \cite{Daelemans+10}.\footnotemark[1]

Using IB1-IG or IGTree for language modeling was proposed by \citet{VandenBosch05b} and further developed by \citet{VandenBosch06b,VandenBosch+09}. Memory-based language modeling has been applied to spelling and grammar correction and (personalized) text completion \cite{VandenBosch+12,VandenBosch+13,Stoop+14b,Stoop+14,Berck17}.

The relation between $k$-NN and neural language models has been studied occasionally. Notably, \citet{KhandelwalLJZL20} describe "Nearest Neighbor Language Models" as the linear interpolation of a pre-trained neural LM with a $k$-NN classifier operating on a data store representing training data encoded in the LMM's embedding space. One of their findings is that $k$-NN is relatively effective in the long tail, i.e. in bringing forward relatively low-frequency events into the predicted token distribution compared to the pre-trained LLM. We return to this observation in Section~\ref{zipf}.

The topic of memorization capabilities of LLMs, an explicit property of memory-based language models we return to in Section~\ref{memorization}, has mostly been framed from the viewpoint of the adversarial elicitation of recitations of content of which the reproduction violates copyright or privacy laws \citep{Carlini+21,Carlini+23,hayes-etal-2025-measuring}. Memorization in neural LLMs is generally observed in these studies to be hard to trigger, to occur only sparsely, and to require models to be large, inference to be greedy (e.g. by setting the hyperparameter temperature to 0), and data to be frequently present in the training data. 

\subsection{Architecture}

\begin{figure}[htb]
  \includegraphics[width=1.14\columnwidth]{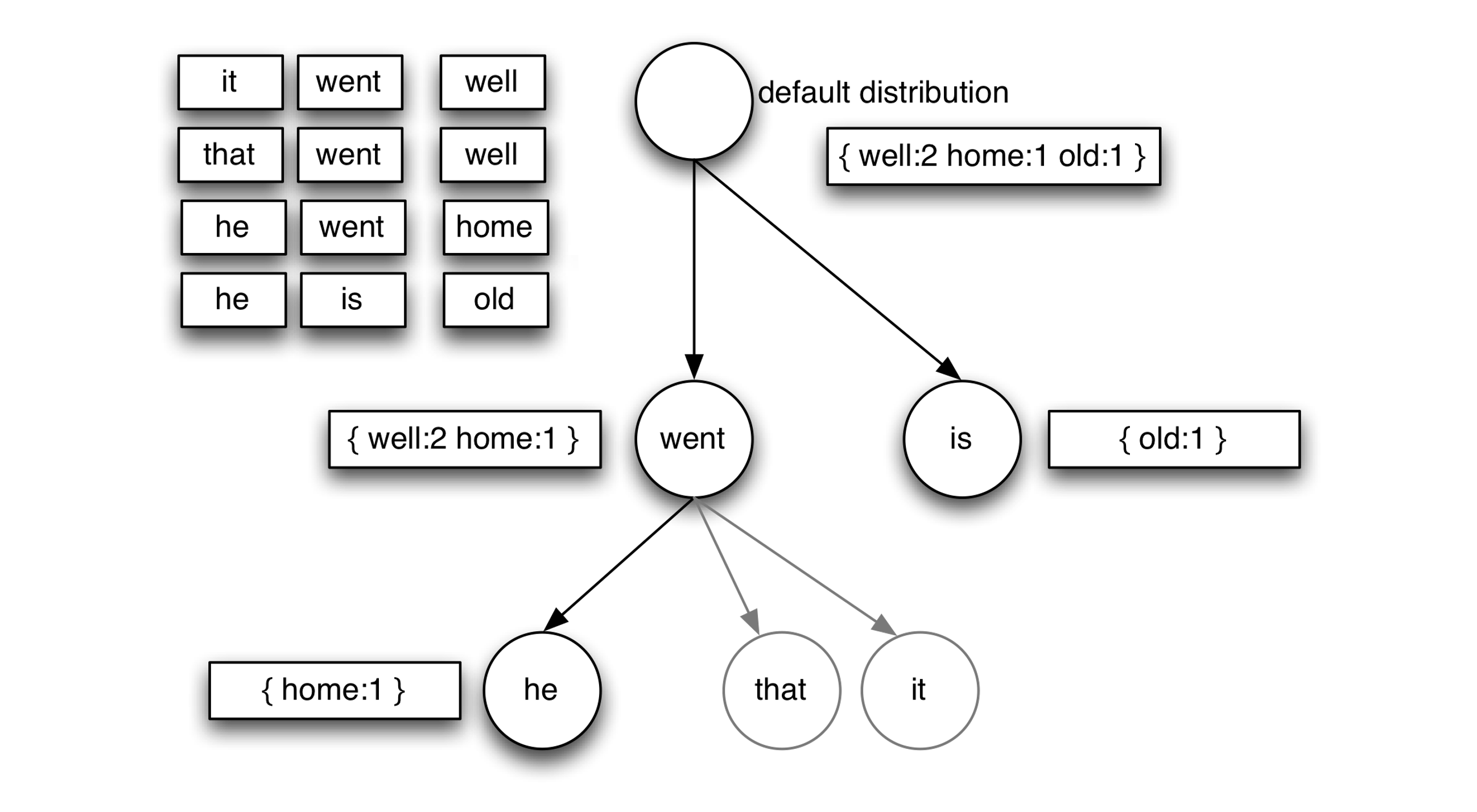}
  \caption{Schematic conversion of an instance base with four examples of two context words predicting a third word, into a prefix trie with next-token distribution information stored at all nodes. Grey nodes, representing subsets of the instance base of which the majority class does not conflict with the parent node, are not stored when the prefix trie is used for decision-tree classification (the IGTree mode of {\sc olifant}).}
  \label{trie}
\end{figure}

A memory-based next-token predictor takes as input a fixed number $n$ of position-specific context tokens positioned before the token to be predicted, $[w_{-n} \ldots w_{-1}]$, and predicts a distribution of $m$ next tokens likely to occur at position $w_0$ ($m \geq 1$), where each candidate is associated with a count (representing the number of neighbors labeled with that candidate). Counts may optionally be softmaxed to sum to $1$.

We have experimented with three variants of memory-based language modeling that differ in how the classifier is trained and how it performs classification or inference \cite{Daelemans+10}. All three make use of the same underlying storage structure, a so-called prefix trie \cite{Knuth73}, offering efficient retrieval of classification information. Figure~\ref{trie} illustrates how a simple training set of four instances is represented in this tree structure. Instances represent contexts of two tokens (e.g. {\em it went}) predicting the third next token ({\em well}). The order in which context words are represented at levels of the tree is automatically determined by computing the gain ratio \cite{Quinlan93} of all context positions. With next-token prediction this amounts invariably to first checking the token immediately before the token to be predicted, $w_{-1}$, followed by the token before that, $w_{-2}$, etcetera. Differing from standard decision-tree classifiers, gain ratio is only computed once for the entire training set for the prefix trie; each level of the trie tests one feature. The layer under the top node tests values occurring in $w_{-1}$; the layer under that tests $w_{-2}$; etcetera.

The top node of the trie represents the full distribution of possible next-token outcomes with absolute counts (optionally normalized to probability distributions). At each next level of the trie, a path is generated for every value that represents a subset of the original instance base in which more than one outcome is still possible.

\begin{figure}[htb]
  \includegraphics[width=\columnwidth]{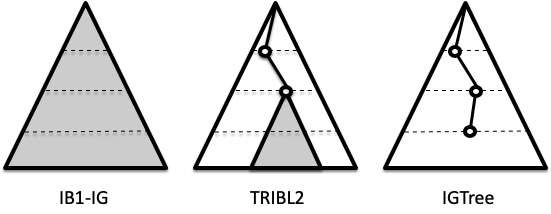}
  \caption{Schematic visualization of classification in the three memory-based language modeling variants. The larger triangle represents the entire prefix trie; grayed zones represent $k$-NN classification; nodes and edges represent downward-pass decision-tree traversal of the prefix trie.}
  \label{memlm-variants}
\end{figure}

Once an instance base is converted to a prefix trie, {\sc olifant} can run in three increasingly faster modes of classification, as also visualized in Figure~\ref{memlm-variants}:

\begin{enumerate}
\item IB1-IG \cite{Aha+91}: a functionally faithful implementation of $k$-NN classification with information-theoretic feature weighting. Based on a new unclassified instance, the $k$ most similar instances are identified in the prefix trie, producing as output the distribution of class labels stored with these $k$ nearest neighbors. The default similarity metric used in all experiments is the Overlap (or L1, or Manhattan) distance weighted per feature by the feature's information gain \cite{Daelemans+10}, as in Equation~\ref{overlap}, where $X$ is the test instance represented by $n$ features, ${x_{0} \ldots x_{n}}$, and $Y$ is every memorized instance represented also by $n$ features. The distance between the two instances, $\Delta$, is a sum of weighted distances per feature, where $\delta$ is the overlap function $\delta(x_{i},y_{i})$, as in Equation~\ref{overlapeq}.

\begin{equation}
\Delta(X,Y) = \sum_{i=1}^{n}\ w_{i} \ \delta(x_{i},y_{i})
\label{overlap}
\end{equation} 

\begin{equation}
\delta(x_{i}, y_{i}) = \left\{ \begin{array}{ll}
                0 & \mbox{if $x_{i} = y_{i}$}\\
                1 & \mbox{if $x_{i} \neq y_{i}$}\\
        \end{array} \right.
\label{overlapeq}
\end{equation}

By default, the weight of each feature is its information gain ratio~\cite{Quinlan93}, as in Equation~\ref{delta}. This is the reduction in the entropy of the classification label, the token to be predicted ($H(C) = - \sum_{c \in C} P(c) \log P(c)$, $C$ representing the next token, $c \in C$ representing all possible next tokens, $V_i$ the set of values at feature $i$ when knowing the value of a feature, divided by the entropy of the feature itself, $si(i)$ as in Equation~\ref{splitinfo}.

\begin{equation}
w_{i} = \frac{H(C) -  \sum_{v \in V_i} P(v) \times H(C|v)}{si(i)}
\label{delta}
\end{equation}

\begin{equation}
si(i) = - \sum_{v \in V_i} P(v) \log P(v)
\label{splitinfo}
\end{equation}

Classification involves multiple traversals of the trie until the $k$ nearest neighbors are identified; the set of equidistant neighbors at the closest distance $\Delta(X,Y)$. This search is somewhat optimized by avoiding the traversal of subtrees in the trie that are guaranteed not to produce nearest neighbors closer than the most distant of the current top-$k$ neighbors, as the sum of remaining potentially matching feature weights is lower than that. In our experiments, we kept $k=1$;

\item TRIBL2 \cite{Daelemans+10} begins the classification process as a decision-tree classifier traversing down the tree in a single path, matching the values of the unclassified instance to values stored in the trie in their gain ratio order. Upon halting at a trie level due to failing to find a match between the currently tested input feature value and values stored at the trie nodes, it switches to perform $k$-NN classification on the remaining sub-instance base under the last matching parent node in the prefix trie. This hybrid offers a trade-off between the sensitivity of IB1-IG (using the same full trie) and the speed of the third option, IGTree;

\item IGTree \cite{Daelemans+97} utilizes the prefix trie fully as a decision tree. Given a new unclassified instance, a single path in the tree is traversed, matching the values of the new instance against nodes in their gain ratio order. Classification occurs when no matching node is found at the next trie level or a matching leaf node is met; the next-token distribution stored at the last matching node is then returned as output. IGTree implements a typical decision-tree "lossy" heuristic, viz. to not store nodes that agree with the majority class label of their parent node (visualized by the grey labels in Figure~\ref{trie}). In our experiments, not storing these nodes (and their sub-trees) typically yields compression rates of about 95\%, compared to losslessly storing all information concerning all training instances (as in IB1-IG and TRIBL2).

\end{enumerate}

\subsection{Scaling dimensions in training}

\begin{figure}[htb]
  \includegraphics[width=\columnwidth]{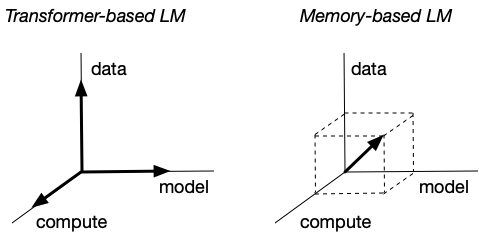}
  \caption{Schematic visualization of the three uncoupled scaling dimensions in training Transformer-based LMs, versus the single coupled scaling dimension involved when training memory-based LMs, where the model equals the data.}
  \label{scaling}
\end{figure}

Before we present empirical scaling results of both training and autoregressive next-token prediction of the three memory-based language modeling variants in Section~\ref{learningcurves}, one difference between scaling in Transformer-based LMs and memory-based LMs should be clarified. Visualized schematically in Figure~\ref{scaling}, training Transformer-based LMs involves making choices in three uncoupled dimensions: the amount of training data, the size of the model, and the amount of compute invested in iterative training. In contrast, in memory-based language modeling, the model is quite literally the data, and the amount of data fully and linearly determines the amount of compute involved in creating the prefix tries.

Due to their complex interactions, studying the cross-effects of scaling in the data, model, and compute dimensions in Transformers is not trivial \cite{journals/corr/abs-2001-08361}. Studying scaling in memory-based LMs, in contrast, can be reduced to learning curve analyses, where training sets are increased and held-out test material is used to measure the effects of more training data.

\section{Learning curve analyses}
\label{learningcurves}

As suggested by \citet{VandenBosch05b}, learning curves measuring next-token prediction accuracy should show that with every $n$-fold increase of the training set, prediction accuracy increases by a roughly constant amount. If this is the case, extrapolations of performance at larger amounts of training data that take larger computational resources than currently available can be made within reasonable bounds of likelihood and precision. 
%In this section we report on a series of learning curve experiments with the three {\sc memlm} variants.

The source for data in all experiments is the EduFineWeb corpus, a freely available 1.3 trillion token corpus consisting of web-crawled data, curated for educational content.\footnote{\url{https://huggingface.co/datasets/HuggingFaceFW/fineweb-edu}} Training data is taken from the first shards of this corpus, up to five billion tokens. The test data consists of the first 10 thousand lines of the first shard of EduFineWeb's validation subcorpus, amounting to 512,660 tokens. Throughout the experiments we used one tokenizer, the GPT2 tokenizer available from Hugging Face\footnote{\url{https://huggingface.co/openai-community/gpt2}} for proper comparison. This tokenizer has 50,257 entries in its vocabulary, composed of 50,000 tokens resulting from byte-level BPE \cite{sennrich-etal-2016-neural}, the 256 base characters as single tokens, and a special end-of-text token.

\subsection{Context width}

\begin{figure}[htb]
  \includegraphics[width=\columnwidth]{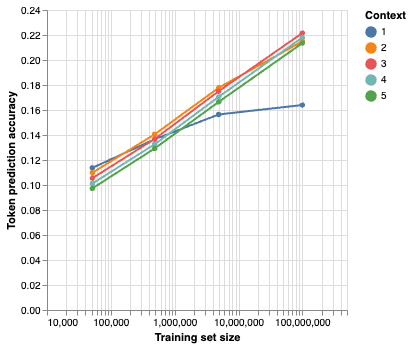}
  \caption{Learning curves in terms of correctly predicted tokens of TRIBL2 with increasing context widths.}
  \label{contexts}
\end{figure}

A first series of learning curve experiments was performed to establish a sufficient context width. With TRIBL2, the middle-ground variant between IB1-IG ($k$--nearest neighbor classification) and IGTree (decision-tree classification), we performed training runs with increasing training set sizes, from 10,000 lines with increments of factor 10, up to 100 million training instances. Checkpointing is available as an option, i.e., trained IB1-IG and TRIBL2 models can be saved and loaded again when training is continued with more data, as $k$-NN models are incremental learners by default, and a prefix trie can be incrementally expanded. In all runs, token prediction accuracy was measured using the same validation set. Figure~\ref{contexts} shows learning curves for context widths $[1, \ldots, 5]$. 

With context width $1$, the next token is predicted based on only the preceding token. In Figure~\ref{contexts}, this blue graph starts with the highest accuracy, but tapers off quickly with more training data. It appears from this graph (given the particular training and validation data used) that the correct next token can be predicted based on only the previous token at a ceiling performance of about $16\%$. The graph of context width 2 tapers off at a later point; the accuracies for context widths 3, 4 and 5 seem to continue going up while hardly differing. In the remainder of our experiments, we continue working with a context width of 4 for all memory-based variants. A context of 4 tokens offers substantially less context than the typical context width employed by Transformer-based next-token predictors.

\subsection{Comparison with GPT variants}

We compare our implementation of memory-based LMs, {\sc olifant}, against the freely available GPT-2 family \cite{radford2019language} and EleutherAI's GPT-Neo family \cite{gao2020pile}. Both sets of pre-trained Transformers include models of multiple sizes, trained on the same training data per family. Both models and training data sets are small enough to offer reasonable comparisons with {\sc olifant}, although the resources available to us are substantially smaller.\footnote{For training sets under a billion tokens, training was performed on an AMD EPYC 7313P 16-Core Processor; for larger training sets, training was done on high-memory CPU nodes with 4TB up to 8TB directly addressable RAM, made available by the Netherlands Snellius HPC facility, \url{https://servicedesk.surf.nl/wiki/spaces/WIKI/pages/30660184/Snellius}} GPT-2 comes in {\tt small} (124M parameters), {\tt medium} (355M parameters), {\tt large} (774M parameters), and {\tt xl} (1.5B parameters) variants, and is trained on the unreleased {\em WebText} corpus, a 38GB cleaned web-crawled corpus with an estimated 8,892,000,000 tokens.\footnote{For this estimate we took as a basis an estimated 180 million English words per GB of normal web text, and a factor of 1.3 to go from English text to English tokenized text: $180 \times 38 = 6,840,000,000$ words, $\times 1.3 = 8,892,000,000$ tokens.} GPT-Neo is available in two sizes, 1.3B and 2.7B, and is trained on the highly problematic and unavailable {\em The Pile} dataset, 889GB of raw text from a multitude of sources, with an estimated total of 196,735,500,000 tokens, about two orders of magnitude more than {\em WebText} and three orders of magnitude more than our current maximal training set size.

\subsection{Token prediction accuracy}

\begin{figure*}[htb]
  \includegraphics[width=\textwidth]{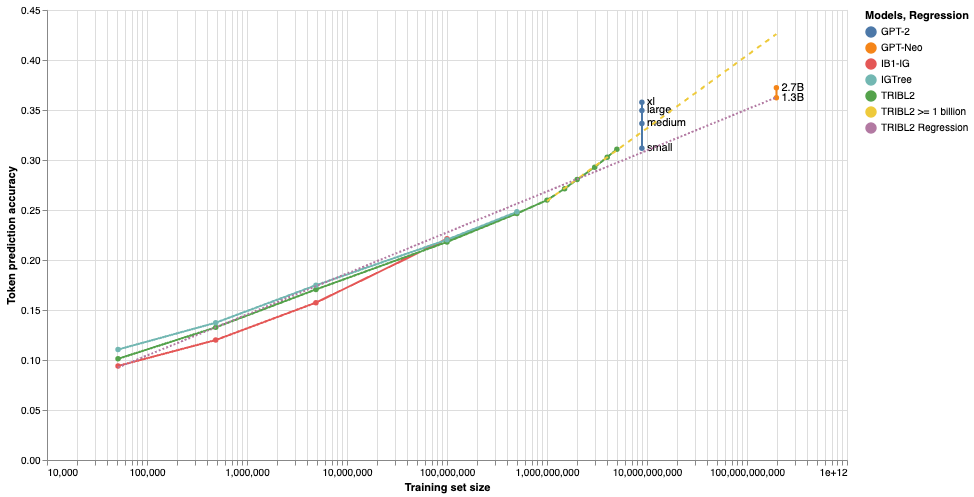}
  \caption{Learning curves on next-token prediction accuracy of the three {\sc olifant} variants and the two GPT systems (with fixed training data sets but increasing model sizes). Dashed and dotted lines are regression functions fitted to all TRIBL2 measurements (dotted) and measurements from 1 billion training tokens (dashed).}
  \label{learningcurve-memlm}
\end{figure*}

Figure~\ref{learningcurve-memlm} shows the percentage of correctly predicted tokens at various training set sizes of the three {\sc olifant} variants, compared against the GPT systems. The GPT variants have fixed training set sizes, hence their 'vertical' graphs. Unlike the memory-based variants, they have been trained iteratively over multiple epochs on their respective training sets. 

Surprisingly, the performance graph of TRIBL2\footnote{At the time of writing, results for IGTree beyond 100 million token training sets were not available yet.} does not exactly show a log-linear relation between training set size and token prediction accuracy, as observed earlier by \citet{van-den-bosch-2006-word} with smaller training set sizes. There seems to be a bend in the curve at around 1 billion training tokens; after the bend, the slope of performance increase is steeper than before this point. First, we performed a log-linear regression analysis on the whole range of measurements for TRIBL2 (the dotted line in Figure~\ref{learningcurves}). The  correlation with the actual measurements is high, with $r = 0.9935$. The regression function, $Token\ prediction\ accuracy = -0.1006 + 0.0178 \times ln (Training\ set\ size)$, predicts an  increase in prediction accuracy of $4.1\%$ with every 10-fold increase in training set size. Based on this estimate, if trained on 1 trillion tokens, TRIBL2 would predict an estimated $39.2\%$ of all tokens correctly.

Assuming that a change in performance increase happens around the 1-billion-token training set size, we performed an additional logistic regression on the measurements starting from the 1 billion token training set size upwards. The correlation of this regression with these five  measurements is high ($r = 0.9993$). The regression function, $Token\ prediction\ accuracy = -0.3960 + 0.0316 \times ln (Training\ set\ size)$, now predicts an increase in accuracy of $7.3\%$ with every 10-fold increase in training set size. Extrapolating, trained on 1 trillion tokens TRIBL2 would predict an estimated $47.7\%$ of all tokens correctly. We leave answering the obvious underlying question, what causes this apparent shift in performance slope, for future research.

The remaining prediction accuracy surplus of the GPT systems over the extrapolated performance of {\sc olifant} variants, at the training set sizes of the WebText corpus or The Pile, must be attributed to the benefit of the way the attention mechanism works \cite{10.5555/3295222.3295349} and how it is able to utilize information from a wide context to predict next words more accurately. While {\sc olifant} works only with four context tokens, both GPT-2 and GPT-Neo operate on a buffer of a much wider context of 1,024 tokens.\footnote{Different from GPT-2, GPT-Neo uses local attention in every other layer with a window size of 256 tokens; see \url{https://huggingface.co/docs/transformers/v4.18.0/model_doc/gpt_neo}} On the other hand, the extrapolated regression function suggests that four context tokens offer enough information to approach or even surpass the performance of these older, smaller GPT systems in next-token prediction.

\subsection{Token prediction latency}

\begin{figure*}[htb]
  \includegraphics[width=\textwidth]{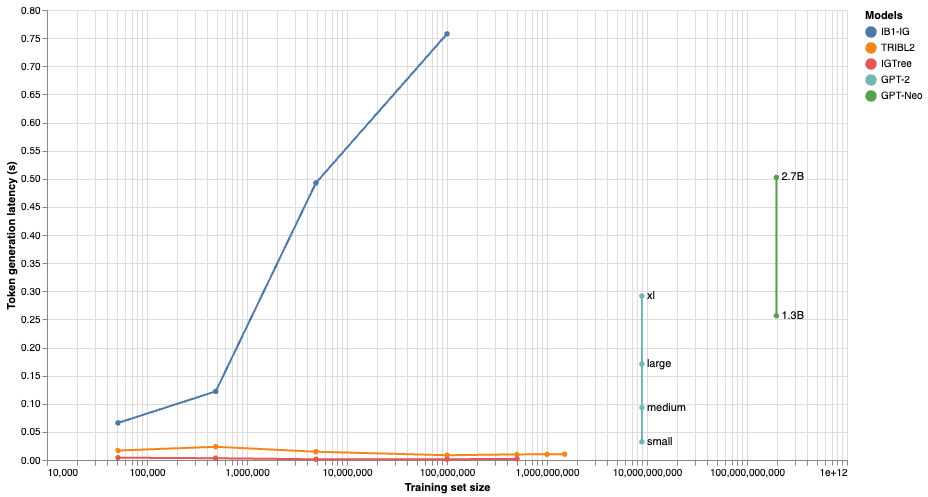}
  \caption{Next-token prediction accuracy latencies (s). Lower is better.}
  \label{latencies}
\end{figure*}

As expected, there is a large difference in the speeds of the three {\sc olifant} variants, despite their similar accuracies. Figure~\ref{latencies} compares the latencies in seconds between the generation of two tokens of the three {\sc olifant} variants and the GPT systems. For comparison, the GPT systems are run on CPUs; in particular, they are run in parallel on 16 CPU cores. 

While IB1-IG turns out very slow, taking nearly a second per prediction when trained on 100 million tokens, the TRIBL2 and IGTree variants exhibit very low latencies, lower than any of the GPT systems. In terms of predicted tokens per second, trained on 100 million tokens, TRIBL2 clocks 118 tokens per second; IGTree performs predictably faster with its single-pass decision-tree classification method, clocking 739 tokens per second, all at roughly the same next-token prediction accuracy.\footnote{With training sets larger than one billion, our measurements show a variance that is due to differences in hardware that was beyond our control. To save electricity, we have restricted ourselves to single runs.}

\subsection{Estimated CO$_2$ emissions}

\begin{figure*}[htb]
  \includegraphics[width=\textwidth]{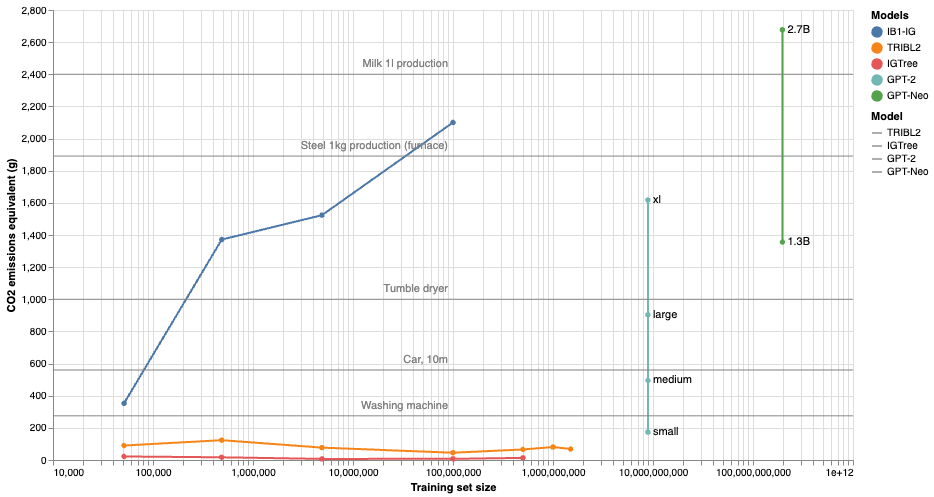}
  \caption{CO$_2$ emission equivalents (g) of the various models on predicting the EduFineWeb validation set of 10,000 lines (512,660 tokens). Grey horizontal lines represent real-world CO$_2$ emission examples.}
  \label{emissions}
\end{figure*}

CO$_2$ emissions during inference, estimated from electricity usage and carbon intensity estimates, roughly follow the same pattern as the latency analysis of the previous subsection, as slower performance implies a longer, thus higher electricity use. Figure~\ref{emissions} displays the CO$_2$ emission equivalent estimations during the full processing by all compared models of the full EduFineWeb validation set (10 thousand lines, 512,660 tokens) as computed by CodeCarbon\footnote{We used CodeCarbon 3.0.0; see \url{https://codecarbon.io/}}, which collects live electricity usage and power measurements from the BIOS of the computer the experiment is run on. Measurements of kWh usage of RAM and CPU were added to arrive at a total usage per experiment. As experiments were run on a stand-alone server (not performing any other tasks) located in Germany, a carbon intensity of $344$ is used.\footnote{The carbon intensity estimate for Germany is for 2024; see \url{https://ourworldindata.org/grapher/carbon-intensity-electricity?mapSelect=~DEU}} As with the latency measurements, we ran the GPT systems on CPU, rather than GPU, for a directly comparable measurement. When instead run on a T4 GPU, GPT-2 small processes the entire validation set about three times faster than when run on CPU, costing about three times less electricity. This illustrates that in contrast to training, where GPUs offer orders of magnitude faster processing, the intrinsically serial process of autoregressive inference may actually be in the same order of magnitude on CPUs as they are on GPUs with neural LMs.

The figure also adds common CO$_2$ emission equivalents for a single washing machine run, driving a car for 10 minutes, a tumble dryer run, and the emissions involved with the furnace-based production of $1 kg$ of steel, and the aggregate emissions involved in the production of $1 l$ of cow milk.\footnote{Statistics taken from \url{https://clevercarbon.io/carbon-footprint-of-common-items}}

From Figure~\ref{emissions} we can observe that while IB1-IG scales rather unfavorably when trained on more data, both its approximations TRIBL2 and IGTree show low CO$_2$ equivalent emissions during inference, regardless of training set size. The internal prefix trie structure causes retrieval to be efficient, staying well below 200 grams of CO$_2$ equivalents when processing the entire validation set. Looking closer at the results for TRIBL2 and IGTree, they show some non-monotonic variance and some larger variability with the largest training set sizes. In contrast, the largest of the small GPT systems in our comparison, GPT-Neo2.7B, emits about 2.7 kilograms of CO$_2$ when processing the validation set.\footnote{Accidentally, the size of each GPT system in billions of parameters is similar to its CO$_2$ emission equivalent in kg when processing this validation set.}

Table~\ref{co2pertoken} lists the CO$_2$ emissions factored to the token level for TRIBL2 and IGTree trained on 500 million tokens, and the largest GPT2 and GPT-Neo models. Set in milligrams, they underscore again the large difference in footprint between the two architectures.\footnote{The numbers also suggest that GPT-Neo 2.7B can generate about 191 tokens within one gram CO$_2$ emission equivalent; this is reminiscent of the various estimates of larger GPT systems emitting 2 to 5 grams per query (often, amortized emissions incorporating the cost of training are factored into these estimates). By comparison, IGTree generates only 5.2 mg of CO$_2$ emission equivalent when generating 200 tokens.}

\begin{table}[htb]
  \centering
  \begin{tabular}{lr}
    \hline
                               & \textbf{CO$_2$ emission eq.} \\
    \textbf{Model}             & \textbf{mg per token} \\
    \hline
    TRIBL2 500m     & 0.128 \\
    IGTree 500m     & 0.026 \\
    \hline
    GPT-2 XL 1.5B   & 3.155 \\
    GPT-Neo 2.7B    & 5.221 \\
    \hline
  \end{tabular}
  \caption{\label{co2pertoken}
    A comparison of CO$_2$ emission equivalents in mg per token during next-token prediction.
  }
\end{table}

\subsection{Model sizes and CO$_2$ emissions during training}

Training {\sc olifant} systems consists of reading the training data, converting it to a prefix trie, and saving this structure to file. Linear in the number of training instances, training can take multiple hours to days with the larger training sets. Using the CodeCarbon package we logged electricity usage during the training of the TRIBL2 and IGTree systems; the estimated CO$_2$ emission equivalents are listed in Table~\ref{training}. They capture the entire training procedure, from reading the training data to saving the prefix trie. They also show the time used for training. The training times for IGTree and TRIBL2 are comparable but differ slightly. In general, IGTree spends extra time on pruning the prefix trie, but since it is 90\%--95\% smaller, it takes less time to write to file. 

\begin{table}[htb]
  \centering
  \begin{tabular}{lrrrr}
    \hline
                   & \multicolumn{4}{c}{\textbf{time (mm:ss) and CO$_2$}} \\
                   & \multicolumn{4}{c}{\textbf{emission eq. (g) during training on}} \\
                   & \multicolumn{2}{c}{4,890,203} & \multicolumn{2}{c}{100,000,000} \\
    \textbf{Model} & time & g CO$_2$ & time & g CO$_2$ \\
    \hline
    TRIBL2         & 01:45 & 0.51 & 33:23 & 13.7 \\
    IGTree         & 01:42 & 0.50 & 34:43 & 14.0 \\
    \hline
  \end{tabular}
  \caption{\label{training}
    Training timings and CO$_2$ emission equivalents (g) with two training set sizes for TRIBL2 and IGTree.
  }
\end{table}

For comparison, the Transformer-based BLOOM language model\footnote{\url{https://huggingface.co/bigscience/bloom}} of 176 billion parameters was trained on 350 billion tokens, and is estimated to have cost 24.7 tonnes (24,700 kg) of CO$_2$ emissions to train \cite{10.5555/3648699.3648952}. If TRIBL2 and IGTree are assumed to consume electricity at about 140 grams per billion training tokens, it would cost the equivalent of about 49 kilograms of CO$_2$ to train on the same number of training tokens as BLOOM 176B, three orders of magnitude less.\footnote{At the average European level, a new passenger car produces 93.6 grams of CO$_2$ per kilometer (\url{https://www.eea.europa.eu/en/analysis/indicators/co2-performance-of-new-passenger}); 49 kilograms of CO$_2$ is produced when the average new car drives 523 kilometers. It takes 263,889 kilometers for the same car to reach the equivalent emissions of training BLOOM 176B. The paper reporting on BLOOM also makes a guess on the CO$_2$ emissions of training GPT-3: 502 tonnes.}

\begin{figure}[htb]
  \includegraphics[width=\columnwidth]{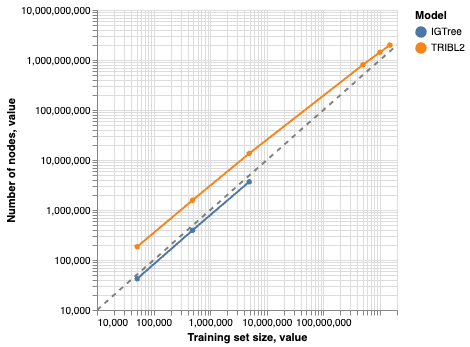}
  \caption{Tree size (in nodes) of IGTree and TRIBL2 as a function of training set size. The dashed grey line represents the parity $x=y$ line.}
  \label{treesize}
\end{figure}

Model sizes also scale linearly with the amount of training data. Figure~\ref{treesize} compares the tree sizes (in nodes) of IGTree and TRIBL trained with context width 4, with increasing training set sizes. Both axes of the figure are logarithmic. The graphs show a structural advantage of IGTree over TRIBL2 in terms of tree size, with IGTree operating on about 90\% to 95\% (one order of magnitude) smaller trees,  when trained on the same amount of data. 

TiMBL, the software engine underlying our implementation, allocates 40 bytes to a single tree node. The TRIBL2 model trained on 1 billion tokens is a tree with 1,428,959,863 nodes: each example with four context tokens contributes just over 1.4 nodes on average. Storing this tree in memory costs 57,158,394,520 bytes, which is just over 53.23 gigabytes. When trained on 2 billion tokens, the tree has 2,507,369,723 nodes (a contribution of 1.25 nodes per training example), which costs 93.4 gigabytes of memory; scaling is slightly sub-linear.

\section{Memorization}
\label{memorization}

While neural LMs are hard-pressed to recite materials in their training data for long stretches of tokens~\cite{Carlini+21}, especially with the temperature hyperparameter set higher than $0$, memory-based LMs are able to reproduce their training data up to a high level of accuracy. TRIBL2 should do this most faithfully as it is a $k$-NN classifier (with fast indexing), while the IGTree variants may be somewhat less precise due to their reliance on decision-tree classification in a pruned prefix trie. Table~\ref{tab:memorization} shows the memorization capacities of both variants trained with context width 4, tested token by token on the first 10 thousand lines (486,724 tokens) of the first shard of the EduFineWeb corpus which they were also trained on. 

\begin{table}[htb]
  \centering
  \begin{tabular}{lrrr}
    \hline
                   & \multicolumn{3}{c}{\textbf{Token memorization (\%) trained on}} \\
    \textbf{Model} &   486,655 & 4,890,203 & 100,000,000 \\
    \hline
    TRIBL2         &     96.32 &     92.71 &       82.52 \\
    IGTree         &     93.14 &     89.62 &       80.02 \\
    \hline
  \end{tabular}
  \caption{\label{tab:memorization}
    Memorization accuracy at the token level (\%) of TRIBL2 and IGTree tested on the first 10 thousand lines (486,655 tokens) of the training set, when trained on this exact same data set and two larger training sets, both including the first 10 thousand lines.
  }
\end{table}

The first observation is that memorization accuracies are high, but under 100\%. The expected best memorizer, TRIBL2, mispredicts 3.7\% of all memorizations when trained on only the test set itself. A misprediction must be due to ambiguous outputs given an exact match on the context of four previous words. Given two or more exact-matching nearest neighbors predicting different next words, the classifier picks the most likely (i.e. frequent) outcome, which might be different from the actual outcome. If the matches are equally likely (when they have the same occurrence count), the tie is broken randomly, which may also lead to a contrasting outcome.\footnote{Random tie breaking in TRIBL2 and IB1-IG has to be explicitly invoked with the {\tt -R <seed>} argument, otherwise another deterministic procedure is followed \cite{Daelemans+10}. We set {\tt -R 42}.}

A second observation is that when the training material on which the memorization test is performed becomes part of a larger training set, as shown in the two final columns of Table~\ref{tab:memorization}, even more alternative outcomes than the next word in the training data become more likely, overruling the single case in memory; again, given the limited context width of 4 previous words. In the case where the training set contains 100,000,000 instances (the table's rightmost column), the first 10 thousand lines representing only 0.5\% of that training set, close to 20\% of all reproductions of tokens in those first 10 thousand lines are incorrect because the training data contains cases of exactly matching contexts that predict other outcomes as more likely. In sum, memory-based LMs are reasonable but imperfect memorizers of training data, and less so when trained on more data. Also, a context of 4 words is not enough to disambiguate in about one in five predictions.

\section{Explainability}
\label{explainability}

\begin{figure}[htb]
  \includegraphics[width=\columnwidth]{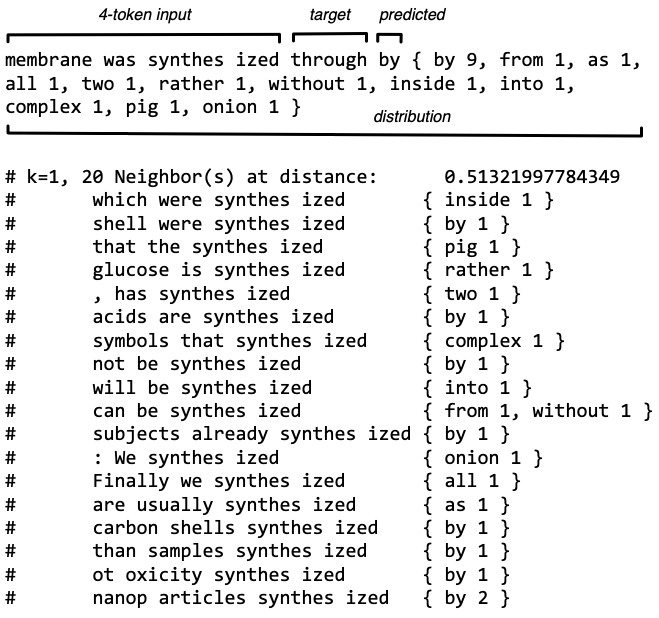}
  \caption{Example output of TRIBL2 classifying a single instance and producing a most likely prediction and a distribution of predictions, based on 20 equally distant nearest neighbors. Token information on word-initial status is left out for legibility.}
  \label{timbloutput}
\end{figure}

Each individual classification of a next word is based on a match in a prefix trie, producing a distribution of possible next words, with (optionally normalized) counts. The TiMBL software allows for various levels of verbosity; depending on the choice of algorithm, it may produce the nearest neighbors found at various distances for a single classification. For instance, Figure~\ref{timbloutput} shows one example classification of predicting the next word in the four-token sequence "[~membrane] [~was] [~synthes] [ized]" (the brackets indicate whether the token is word-initial by including a space in front or not). The next token should be "[~through]".

The predicted next word is, incorrectly, "[~by]". As the output shows, this prediction is by far the most frequent outcome among the 20 nearest neighbors found at the same distance of two matching tokens (the two tokens immediately preceding the predicted token) and two mismatching tokens (the two tokens before that). The correct word "[~through]" is not among the possibilities. In TiMBL, the $k=1$ setting is implemented such that all equidistant neighbors at the closest distance, i.e. with the smallest mismatching number of words weighted by their gain ratio, are taken together as the basis for the output distribution. In other words, in this case the distance is the sum of gain ratio weights of the mismatching features, the two leftmost context tokens; neither "[ membrane]" nor "[ was]" was matched in any closer neighbor. Within the set, two identical neighbors are found with different outcomes ("[~from]" and "[~without]"), and another pair of identical neighbors are found with the same outcome ("[~by]"). All other neighbors represent unique contexts occurring once in the training data.

This example illustrates that prediction by TRIBL2 and IB1-IG can be traced to individual instances stored in the prefix trie. Unique identifiers that would link an instance to its original document could be included in a "hidden" lowest level in the prefix trie so that individual predictions could even be attributed to (locations in) specific sources. Classification in IGTree, however, offers less verbosity; due to its pruning of the prefix trie only the next-word distribution and the depth of the path can be shown.

\section{Perplexity and a distributional analysis of {\sc olifant} predictions}
\label{zipf}

In contrast to largely non-zero softmaxed logits produced over the entire token vocabulary by GPT systems, next-token predictions generated by memory-based models are sparse; by comparison they are almost empty. As an example, TRIBL2 trained on 100 million tokens predicts next tokens based on nearest-neighbor distributions with a median of only 6 tokens, from which the most likely next token is selected. As we set $k=1$ throughout our experiments, most IB1-IG and many TRIBL2 predictions are based on a single nearest neighbor, or on a set of equidistant nearest neighbors. Larger distributions do occur, however; the mean number of next-token predictions in TRIBL2 output is 87.5. IGTree bases its predictions on somewhat larger distributions: its predictions have a median of 13 and a mean of 356 possible next tokens. This difference can be explained by IGTree's earlier stopping at higher levels and in non-ending nodes in the tree, where nodes represent relatively larger subsets of memorized instances.

\begin{figure*}[htb]
  \includegraphics[width=\textwidth]{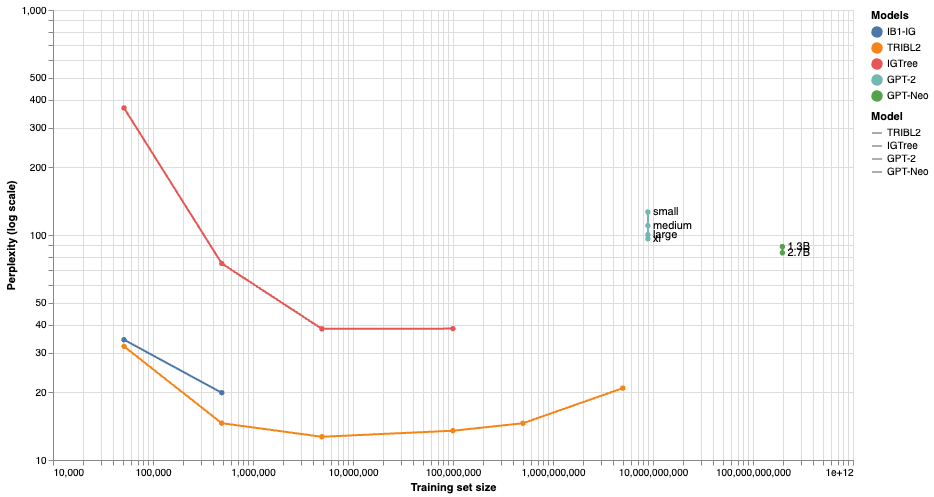}
  \caption{Perplexity learning curves of the different systems. Both axes are logarithmic.}
  \label{perplexity}
\end{figure*}

We computed perplexity scores as the sum of base-2 logprobs assigned by the different systems to the target tokens, averaged over all tokens that received a non-zero probability from the same edufineweb validation set as used before in the learning curve analyses. Using double logarithmic axes, Figure~\ref{perplexity} shows the perplexities attained by the different {\sc olifant} systems trained on increasing amounts of training data, compared to those of GPT-2 and GPT-Neo at different model sizes (and fixed training set sizes). 

From the figure it is apparent that the perplexity scores of IB1-IG and TRIBL2 are considerably lower than those of IGTree and the GPT systems. Zooming in, it turns out that IB1-IG and TRIBL2 frequently assign a zero probability to the actual next token, meaning that we did not include the $-\inf$ logprob in computing perplexity. Trained at 100 million training tokens, IGTree assigns 49.6\% of all actual next tokens a non-zero probability, TRIBL2 assigns this to 41.9\% of all next tokens; trained on 5.5 billion tokens, this rises to 63\%. GPT2-small, to confirm, assigns 100\% of all next tokens a non-zero weight. In other words, IGTree and the GPT-2 systems spread their chances over many if not all possible outcomes, while TRIBL2 and IB1-IG limit this spreading to just a few candidates, often assigning zero probability to the actual next token. Having placed bets on a small number of proverbial horses, TRIBL2 and IB1-IG are less perplex when one of their few best guesses turns out to be the actual next token.

\begin{figure}[htb]
  \includegraphics[width=\columnwidth]{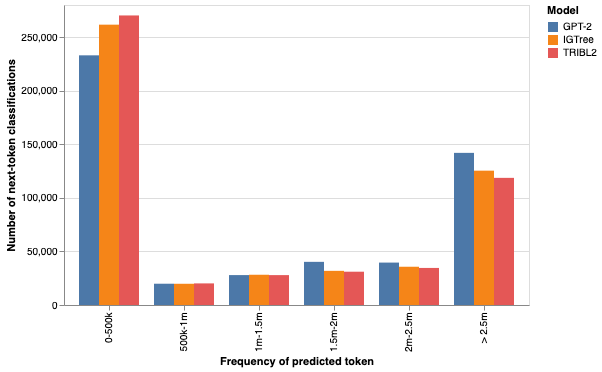}
  \caption{Bar-graph visualization of the distribution of next-token prediction frequencies of occurrence in a 100 million-token training set by TRIBL2 and IGTree, both trained on 100 million instances, and by GPT2-XL, all tested on the fixed validation set.}
  \label{frequencies}
\end{figure}

A difference between TRIBL2 (and IB1-IG) on the one hand, and IGTree on the other hand is which tokens they predict. The token vocabulary used throughout this paper is the one used for GPT-2. Given the counts of all tokens occurring in the 100 million-token training set, Figure~\ref{frequencies} shows the distribution of the binned frequency-of-occurrence counts of the tokens actually predicted by TRIBL2 and IGTree. For comparison, the figure also shows the same numbers for the tokens actually predicted by GPT2-small, which is trained on a different (and larger) training set, the aforementioned 8.9 billion token WebText corpus, using the same tokenizer. The largest three bars on the left represent the tail of the frequency band with low-frequency tokens, and the large bars on the right end represent the head, i.e. the most frequent tokens. Although the differences are less outspoken than in Figure~\ref{perplexity}, it can be observed that GPT2-small predicts fewer low-frequency and more high-frequency tokens than the memory-based models. This confirms the aforementioned observation made by \citet{KhandelwalLJZL20} that $k$-NN is effective in pushing relatively low-frequency events into the predicted token distribution compared to neural language models.

\section{Discussion and conclusions}
\label{conclusions}

Although the best performances presented in this paper, obtained in the fall of 2025, are only approaching those of GPT-2 from six years ago, we argue that memory-based language models offer an interesting alternative to neural foundation models. They may prove useful in contexts where relatively small and fast foundation models are needed as a component, e.g. in speculative decoding \cite{10.5555/3618408.3619203}. Trained on more data, their performance can reliably predicted to go up. Our learning curve study even showed an increase in the steepness of log-linear performance growth beyond one billion training tokens; yet, we cannot be certain about the curves flattening at a later point, e.g. because of the inherent limitations of the four-token context we have now restricted our models to. 

We believe from our emission estimates that we argue that our models' small ecological footprint could be an advantage in many situations, also at scale. {\sc olifant}'s reliance on CPUs rather than GPUs or TPUs opens alternative paths to scaling with potentially drastic ramifications on electricity usage, and also on costs.

Summarizing, our results and analyses have highlighted favorable properties of {\sc olifant}'s fast approximations of $k$-nearest neighbor classification, TRIBL2 and IGTree, with TRIBL2 having a few more advantages over the faster of the two, IGTree:

\begin{description}
    \item[Learning is scalable, and incremental with TRIBL2.] Model performance increases approximately log-linearly with more data, or better; model size scales linearly. IGTree produces 90\%-95\% smaller models than TRIBL2 and processes in the order of hundreds of tokens per second. TRIBL2 is a natural incremental learner and can learn and finetune with checkpointing;
    \item[Consistently low CO$_2$ emissions] during training and inference. Our packaged implementation of memory-based language models {\sc olifant} runs on CPUs. Using extrapolations and reported estimates for Transformer-based systems, training with {\sc olifant} is estimated to cause in the order of 1,000 times fewer CO$_2$ emissions than neural LM training. Inference costs are 10-100 times lower;
    \item[Fully transparent functioning.] TRIBL2 offers nearest-neighbor-based explanations for predictions, based on individual examples used for prediction. Nearest neighbors can even be source-tagged for full provenance;
    \item[Intentional memorization of training data.] At context size 4, {\sc olifant} models can recite 80\% or more of tokens from their training data faithfully.
\end{description}

For completeness we note that {\sc olifant}'s underlying engine, TiMBL, can be parallelized on multi-core CPUs both for training \cite{VandenBosch+07b} and inference. For inference TiMBL can run in multiprocessor mode, effectively cloning itself \cite{Daelemans+10}. Depending on the way TiMBL is integrated in other software,\footnote{TiMBL can run as a server (\url{https://github.com/LanguageMachines/timblserver}); a TiMBL API is available, as well as TiMBL Python bindings (\url{https://github.com/proycon/python-timbl}).} the availability of $n$ CPU cores can potentially lead to a speedup of up to a factor $n$ in training, in inference and in specific settings such as beam search or speculative decoding.

An obvious next step is to continue incremental training, and to perform benchmarking on standard tasks for decoders, starting with older benchmarks such as GLUE \cite{wang-etal-2018-glue}. That said, we would like to argue that next-token prediction is a valid benchmark task in its own right, that brings to light differences at small training sets, at which performance on more complex benchmark tasks would still be at baseline level.

\subsection*{Limitations and future work}

The key current limitation of the memory-based language modeling approach as implemented in {\sc olifant} is that it is now confined to a position-specific context with a microscopically small width compared to the usual neural LM context width. We have experimented with larger position-specific contexts, up to 16 words. For IGTree, wider contexts lead to larger tries but hardly better predictions; for TRIBL2, wider contexts lead to larger tries, slower performance and small gains in next-token prediction accuracies. Context information with predictive value that is obviously available in the wider context, but that arguably should not be encoded in position-specific features, could be made available in other ways, e.g. through encoding it as a bag-of-words vector. This is an obvious avenue for future work.

TiMBL, the underlying engine, offers a number of extensions that could be profitable for generalization performance and could enhance the functionality of {\sc olifant}. First, individual examples can be weighted with an exemplar weight, giving them a weighted vote in $k$-NN classification rather than their count in memory. A prompt or a training corpus presented for finetuning could be incrementally added to memory as they are presented as input, and be given a higher weight that would prioritize their use in the autoregressive generation of new output. This way exemplar weights could function as a kind of attention weights; lowering exemplar weights could emulate a kind of forgetting. 

Second, preliminary experiments have shown a small profitable effect of raising $k > 1$ in TRIBL2 (note again that we have used $k = 1$ throughout the experiments reported here), and using distance-weighting of neighbors \cite{Daelemans+10}, more distant neighbors getting a lower vote in classification. Raining $k$ would in theory increase the size of the non-zero distribution of possible next tokens. Speeds of TRIBL2 will only be slightly affected with increased $k$, as the search for the top-$k$ neighbors is an efficient process that is only moderately influenced by the actual value of $k$. 

Third, TiMBL offers several value-difference metrics, such as MVDM \cite{Cost+93}, that offer more nuanced types of similarity metrics than the Overlap function used here. Acting more or less as static embeddings, MVDM could represent meaningful (e.g. semantic) similarities between two tokens occurring in the same position that could help establishing a closer similarity of more relevant nearest neighbors.

Finally, we note that for a broad uptake of memory-based language modeling, it would be good if it would conform to standards organically evolved by communities on platforms such as Hugging Face,\footnote{\url{https://huggingface.co/}} or if these platforms become more eclectic in the LLM engines they cater for. This would also ease the integration of {\sc olifant} models in larger setups such as speculative decoding \cite{10.5555/3618408.3619203}, and connecting the models to chat and instruct modules. Even though the prediction distributions of IGTree and TRIBL2 are sparse, beam search can be readily applied.

\section*{Acknowledgments}

The authors thank Ko van der Sloot for his sustained and essential contributions to the development of TiMBL.  Walter Daelemans, Ton Weij\-ters, and Jakub Zavrel all provided elemental contributions to the development of the TiMBL family of algorithms; thank you. We also thank Wessel Stoop, Herman Stehouwer, Menno van Zaanen, Tanja Gaustad--van Zaanen, and Monica Hajek for contributions to Valkuil.net, Fowlt.net, and WOPR, predecessor versions of {\sc memlm}, and to Sander Canisius for the first versions of the Python-TiMBL bindings.

\bibliography{ilk,custom}

%\appendix

%\section{Example Appendix}
%\label{sec:appendix}

%This is an appendix.

\end{document}